\documentclass{article} 
\usepackage{nips15submit_e,times}
\usepackage{amsmath,amsfonts,amsthm,amssymb}
\usepackage{hyperref}
\usepackage{url}
\usepackage{graphicx}
\usepackage{epstopdf}

\usepackage{algorithm}
\usepackage{algorithmic}

\usepackage[colorinlistoftodos,bordercolor=orange,backgroundcolor=orange!20,linecolor=orange,textsize=scriptsize]{todonotes}

\newtheorem{theorem}{Theorem}
\newtheorem{definition}[theorem]{Definition}
\newtheorem{assumption}[theorem]{Assumption}

\newtheorem{remark}[theorem]{Remark}

\def\spvecA#1;{\if;#1;\else #1\cr \expandafter \spvecA \fi}

\newcommand{\diag}{{\bf  diag}}
\newcommand{\vsubset}[2]{{#1_{[#2]}}}
\newcommand{\R}{{\mathbb R}}
\newcommand{\Exp}{\mathbf{E}}
\newcommand{\q}{ {\bf q}}

\newcommand{\PP}{\mathbf{P}}
\newcommand{\vc}[2]{#1^{(#2)}}
\newcommand{\vi}[2]{#1_{#2}}
\newcommand{\ve}[2]{\left\langle #1 ,  #2 \right\rangle}
\newcommand{\eqdef}{:=}

\newcommand{\n}{\nabla}

\newcommand{\proj}[2]{{\bf Proj}^W_{#1}\left(#2\right)}
\newcommand{\pproj}[2]{{\bf Proj}_{#1}\left(#2\right)}

\title{Linear Convergence of the Randomized Feasible Descent Method Under the Weak Strong Convexity Assumption}

\author{
Chenxin  Ma
\\
Industrial and Systems Engineering\\ Lehigh University\\
Bethlehem, PA 18015, USA\\
\texttt{chm514@lehigh.edu} \\
\And
Rachael Tappenden \\
 Applied Mathematics and Statistics \\
 Johns Hopkins University \\
Baltimore, MD 21218, USA \\
\texttt{rtappen1@jhu.edu} \\
\AND
Martin  Tak\'a\v{c} \\
Industrial and Systems Engineering\\ Lehigh University\\
Bethlehem, PA 18015, USA\\
\texttt{Takac.MT@gmail.com}
}

 \nipsfinalcopy 
\newcommand\tagthis{\addtocounter{equation}{1}\tag{\theequation}}

\begin{document}

\maketitle

\begin{abstract}
In this paper we generalize the framework of the feasible descent method (FDM) to a randomized (R-FDM) and a coordinate-wise random feasible descent method (RC-FDM) framework. We show that the famous SDCA algorithm for optimizing the SVM dual problem, or the stochastic coordinate descent method for the LASSO problem,
fits into the framework of RC-FDM.
We prove linear convergence for both R-FDM and RC-FDM under the weak strong convexity assumption.
Moreover, we show that the duality gap converges linearly for RC-FDM, which implies that the duality gap also converges linearly for SDCA applied to the SVM dual problem.
\end{abstract}

\section{Introduction}

In this paper we are interested in the following optimization problem
\begin{equation}\label{eq:P}
\min_{x \in X} f(x),
\end{equation}
where the function $f$ is smooth  and convex, and $X\subseteq \R^n$ is a convex set.
The Feasible Descent Method (FDM)
\cite{luo1993error,necoara2015linear,wang2014iteration}
is any algorithm, which produces a sequence of points $\{\vi{x}{k}\}_{k=0}^\infty$, where there exist constants $\beta\geq 0$, $\zeta > 0$ and $\omega_k \geq \bar \omega >0$, such that the following 3 inequalities hold for every iteration $k$:
\begin{align}
\label{FDM1}
\vi{x}{k+1}
 &= \pproj{X}{
 \vi{x}{k} - \vi{\omega}{k}
   \nabla f(\vi{x}{k})
   + \vi{z}{k}
  },\\
  \label{FDM2}
  \|\vi{z}{k}\|
  &\leq \beta   \|\vi{x}{k}
    -\vi{x}{k+1}\|,
    \\
    \label{FDM3}
 f(\vi{x}{k+1})
 &\leq f(\vi{x}{k})
  - \zeta   \|\vi{x}{k} - \vi{x}{k+1}\|^2,
\end{align}
where $\pproj{X}{y}\eqdef \arg\min_{x \in X} \|x-y\|$ is the projection of $y$ onto $X$.

As was shown in \cite{luo1993error}, many first order algorithms,
including steepest descent, the gradient projection algorithm, the extra gradient method, the proximal minimization algorithm and the cyclic coordinate descent method, fit into the framework of FDM. However, randomized first order algorithms are becoming more and more popular nowadays,
and the following question naturally arises:
\begin{center}\it  ``Can the framework of FDM be extended to a randomized setting?"
\end{center}

In this paper we give an affirmative answer to this question: we show that, indeed, a randomized version of FDM can be formulated and we will show that, for example,
the inexact gradient projection algorithm (when the gradient is corrupted with random noise) or the stochastic coordinate descent method, fit into this new framework.

\subsection{Assumptions and Notations}
\label{sec:assumptions}

In this section we state the assumptions and introduce the notation that will be used in this paper.

The first assumption we make is that the function $f$ enjoys weak strong convexity, which is captured by the following.
\begin{assumption}
\label{asm:wsc}
We assume that
there exists a positive vector
$w \in \R^n_{++}$ such that
the function $f(x)$
satisfies the {\em weak strong convexity property on the set $X$}, which is defined as
\begin{equation}
\label{eq:WSC}
f(x) - f(\bar x) \geq \kappa_f \|x-\bar x\|^2_W,\quad \forall x\in X,
\end{equation}
where
$f^* = \arg\min_{x\in X} f(x)$,
$\bar x =
 \arg\min_{y\in X: f(y)=f^*} \|x-y\|_W$, $\| x \|_W^2
 = \sum_{i=1}^n w_i (\vc{x}{i})^2$,
$W =\diag(w)$, and $ \kappa_f >0$.
\end{assumption}

Let us remark that if $f$ is smooth and has a Lipschitz continuous gradient, then Assumption \ref{asm:wsc} is weaker than the {\em strong convexity} assumption or the {\em global error bound property} \cite{necoara2015linear}.

The second assumption we make regards the smoothness of $f$, and is defined precisely as follows.
\begin{assumption}
 \label{asm:lc}

We assume that $f(x)$ has a coordinate-wise Lipschitz continuous gradient with constants $L_i$, i.e.
$\forall x \in X$
and $\forall \delta \in \R: x+\delta e_i \in X$ the following inequality holds
\begin{equation}
\label{eq:CWLC}
 |\nabla_i f(x)
  - \nabla_i f(x+\delta e_i)|
  \leq L_i |\delta|,
\end{equation}
where $e_i$ denotes the $i$-th column of the identity matrix $I \in \R^{n \times n}$.
\end{assumption}
As it was shown in \cite{richtarik2014iteration},
Assumption \ref{asm:lc}
implies that the function $f(x)$
has a Lipschitz continuous gradient with Lipschitz constant $L_f^W > 0$ with respect to the norm $\|\cdot\|_W$, i.e. $\forall x,y\in X$ we have
\begin{equation}
\label{eq:globalLipConstant}
 \|\nabla f(x) - \nabla f(y) \|_W^*
 \leq L_f^W  \|x-y\|_W,
\end{equation}
where $\|x\|_W^*
 = \sqrt{\sum_{i=1}^n \frac1{w_i} (\vc{y}{i})^2}$ is the dual norm to $\|\cdot\|_W$.
 Moreover, it was shown in \cite{richtarik2014iteration} that
 $L_f^W \leq \sum_{i=1}^n \frac{L_i}{w_i}$.

Let us define the projection operator onto the set $X$, with respect to the norm $\|\cdot\|_W$, as follows
\begin{equation}\label{Eq:pd}
\proj{X}{x} = \arg\min_{y \in X} \|x-y\|_W^2
=\arg\min_{y \in X} \sum_{i=1}^n w_i (\vc{x}{i}-\vc{y}{i})^2,
\end{equation}
where $\vc{x}{i}$ denotes the $i$-th coordinate of the vector $x$.

\subsection{Applications}
In this section we discuss several problems that arise in the optimization and machine learning literature, which fit into the FDM framework that we analyze in this paper. We also provide details showing that, for each problem, the objective function satisfies the assumptions in Section \ref{sec:assumptions}. (A discussion on the value of the weak strong convexity parameter $\kappa_f$ will be given in Section \ref{sec:GEB}.)


\paragraph{The dual of SVM.}
Consider the classical linear SVM problem. The goal is, given $n$ training points $(a_i,y_i)$, where $a_i \in \R^d$ are the features for point $i$ and $y_i \in \{-1,+1\}$ is its label, find $w \in \R^d$ such that the regularized empirical loss function is minimized,
i.e., one can minimize the following optimization problem
\begin{equation}\label{eq:prime}
	\min_{w\in \R^d} \left\{ \PP(w) :=\tfrac{1}{n}\textstyle{\sum}_{i=1}^n \ell_i(w^Ta_i) +  \tfrac{\lambda}{2} \|w\|^2  \right\},
\end{equation}
where $\lambda>0$ is a regularization parameter,
and, in the case of SVM, the function
$\ell_i(w^T a_i) = \max\{0,1- y_i w^T a_i\}$ is the hinge loss.
Clearly, the objective function \eqref{eq:prime} is not smooth. However, one can formulate the dual \cite{hsieh2008dual,shalev2013stochastic,TRN:minibatchICML}
\begin{equation}\label{eq:tempdual}
\min_{x \in \R^n, 0\leq \vi{x}{i}\leq 1} \left\{f(x) \eqdef  \tfrac{1}{2\lambda n^2 } x^T Q x- \tfrac{1}{n} {\bf 1}^T x \right\},
\end{equation}
where $Q_{i,j} = y_i y_j \ve{a_i}{a_j}$, and ${\bf 1}$ denotes the vector of all ones, which \emph{is} smooth.

\paragraph{Lasso problem  and least squares problem.}
Consider the following optimization problem
\begin{equation}\label{faw4efewwaef}
\min_{x \in \R^n}
 g(x) + \lambda \|x\|_1,
\end{equation}
where $\lambda \geq 0$ and $g(x)$ is a smooth function with the
special structure:
$g(x) = h(Ax)+q^Tx$, where $A \in \R^{m \times n}$
is some data matrix, $q \in \R^n$ is some vector and
$h$ is a strongly convex function.
It is a simple exercise to show that,
if we double the dimension of $x$ to $[x^+;x^-]$,
we can replace the term $\lambda \|x\|_1$ in \eqref{faw4efewwaef} with
$\lambda {\bf 1}^T x^+ + \lambda {\bf 1}^T x^-$
and impose the constraints $x^+, x^- \geq {\bf 0}$.
Then the Lasso problem \eqref{faw4efewwaef} can be reformulated as a smooth optimization problem with simple box constraints.

 \paragraph{$\ell_2$ regularized empirical loss minimization.}
 Many machine learning problems have the following structure
 \cite{chang2008coordinate}
 $$
 \min f(x) = \frac1n \sum_{i=1}^n \ell_i(a_i^T x) +
 \frac{\lambda}{2} x^Tx,
 $$
 where $\lambda>0$ is a regularization parameter
 and $\ell_i$ is a loss function.
 Because we assume that $f$ must be smooth,
 the following commonly used loss functions fit our assumptions:
 the logistic loss function $\ell_i(a_i^T x) =
 \log(1+\exp(-y_i a_i^Tx))$;
the squared loss function $\ell_i(a_i^T x) =
 (y_i -a_i^Tx)^2$ and
 the squared hinge loss function $\ell_i(a_i^T x) =
 (\max\{0,1-y_i a_i^T x\})^2$.
 \subsection{Related work}

Luo and Tseng \cite{luo1993error} are among the first to establish asymptotic linear convergence for a non-strongly convex problem under the local error bound property. They consider a class of feasible descent methods (which includes e.g. the cyclic coordinate descent method).
The error bound  measures how close the current solution is to the optimal solution set with respect to the projected gradient.
Recently, \cite{wang2014iteration}
proved that the feasible descent method enjoys a linear convergence rate (from the beginning, rather than only locally) under the global error bound property.
Considering the class of smooth constrained optimization problems
with the global error bound property,
 \cite{nec2013,necoara2014distributed} showed a linear convergence rate for the parallel version of the stochastic  coordinate descent method.
In \cite{liu2014asynchronous}
the authors analyzed the asynchronous stochastic coordinate descent method (SCDM) under the weak strong convexity assumption.
Very recently, \cite{necoara2015linear} showed that, if the objective function is smooth, then the class of problems with the global error bound property is a subset of the class of problems with the weak strong convexity property.

\subsection{Contributions}

In this Section we list the most important contributions of this paper (not in order of their significance):
\begin{itemize}
\item
{\bf Randomized and Randomized Coordinate Feasible Descent Methods.}
We extend the well known framework of Feasible Descent Methods (FDM) \cite{luo1993error} to randomized and randomized coordinate FDM and  show that the SCDM  algorithm fits into our new proposed framework.

\item \textbf{Linear Convergence Rate.}
We show that any stochastic or deterministic algorithm, which fits our Randomized FDM (R-FDM) or Randomized Coordinate-FDM (RC-FDM) framework and satisfies our previously stated assumptions, converges linearly in expectation.

\item
{\bf Linear Convergence of the Duality Gap for SDCA for SVM}.
As a consequence of our analysis, we show that when SDCA is applied to the dual of the SVM problem, the duality gap converges linearly.

\end{itemize}

\subsection{Paper Outline}
In Section  \ref{sec:FDM}
we derive the Randomized (R-FDM) and the Randomized Coordinate (RC-FDM) Feasible Descent Method. In Section \ref{sec:CA}
we derive the convergence rate for any method which fits into the R-FDM or RC-FDM framework and we compare our results with those in \cite{liu2014asynchronous} for SCDM.
In Section \ref{sec:GEB} we briefly review the global error bound property and using the result in \cite{necoara2015linear} we compare our convergence results with \cite{wang2014iteration}. In Section \ref{sec:SDCA} we show that the duality gap converges linearly for SDCA applied to the dual of the SVM problem, and in Section~\ref{S_Summary} we present a brief summary.

\section{Randomized and Randomized Coordinate  Feasible Descent Method}
\label{sec:FDM}

The framework of Feasible Descent Methods (FDM) broadly covers many algorithms that use first-order information \cite{luo1993error}
including gradient descent, cyclic coordinate descent and also the inexact gradient descent algorithm.
We generalize the classical FDM framework to a randomized setting, which we call the
Randomized Feasible Descent Method (R-FDM).
To the best of our knowledge this is the first time such a framework has been considered and that a global linear convergence rate has been established under Assumptions
\ref{asm:wsc} and \ref{asm:lc}.
Further, we also show that the popular minibatch stochastic coordinate descent/ascent method fits into the R-FDM framework.

\begin{definition}[Randomized Feasible Descent Method (R-FDM)]\label{Def_RFDM}
A sequence $\{\vi{x}{k}\}_{k=0}^{\infty}$
is
generated by R-FDM if
there exist $\beta\geq0$, $\zeta > 0$
and $\{\vi{\omega}{k}\}_{k=0}^\infty$
with $\min_{k} \vi{\omega}{k}\geq  \bar \omega > 0$
such that
 for every iteration $k$, the following conditions are satisfied
\begin{align}
\label{eq:RFDM-1}
\vi{x}{k+1}
 &= \proj{X}{
 \vi{x}{k} - \vi{\omega}{k}
  W^{-1}
  (\nabla f(\vi{x}{k})
   - \vi{z}{k})
  },\\
  \label{eq:RFDM-2}
\Exp[ (\|\vi{z}{k}\|_W^*)^2]
  &\leq \beta^2 \Exp[\|\vi{x}{k}
    -\vi{x}{k+1}\|_W^2],
    \\
    \label{eq:RFDM-3}
\Exp[ f(\vi{x}{k+1})]
 &\leq f(\vi{x}{k})
  - \zeta  \Exp[\|\vi{x}{k} - \vi{x}{k+1}\|_W^2],
\end{align}
where $\vi{z}{k}$ is some random vector that satisfies the Markov property conditioned on $\vi{x}{k}$.
\end{definition}
We will now compare the new Randomized FDM framework (Definition~\ref{Def_RFDM}) with the original FDM (\eqref{FDM1}--\eqref{FDM3}), where, for simplicity of exposition, we will take $\|\cdot\|_W \equiv \|\cdot\|_2$ (i.e., $W = I$). Notice that the first step of R-FDM \eqref{eq:RFDM-1} is the same as the first step of FDM \eqref{FDM1}. The key difference between FDM and R-FDM is that for FDM, \eqref{FDM2} and \eqref{FDM3} hold deterministically (with a deterministic vector $z_k$), whereas for R-FDM \eqref{FDM2} and \eqref{FDM3} only need to hold \emph{in expectation}. That is, for R-FDM, conditions \eqref{FDM2} and \eqref{FDM3} are \emph{replaced by conditions \eqref{eq:RFDM-2} and \eqref{eq:RFDM-3}}, where $z_k$ is a \emph{random vector}. Notice that \eqref{eq:RFDM-2} and \eqref{eq:RFDM-3} are weaker conditions than \eqref{FDM2} and \eqref{FDM3}. That is, for FDM, \eqref{FDM2} and \eqref{FDM3} must hold at every iteration (i.e., they are deterministic), whereas for the R-FDM framework, the conditions \eqref{eq:RFDM-2} and \eqref{eq:RFDM-3} are equivalent to \eqref{FDM2} and \eqref{FDM3} holding \emph{only on average}. Thus, the R-FDM framework is more general than FDM.

\begin{remark}
We will see later (in the proof of convergence of R-FDM) that \eqref{eq:RFDM-2}
can be relaxed to the existence of constant $\eta>0$ such that
$
\Exp[ (\|\vi{z}{k}\|_W^*)^2]
   \leq
    \eta  \Exp[\|\vi{x}{k} - \vi{x}{k+1}\|_W^2].
  $
\end{remark}

We will now demonstrate that (see Theorem~\ref{thm:sdca_is_RFDM}), under an additional mild assumption, if the set $X = \R^n$, then SCDM (captured in Algorithm \ref{alg:sdca} with Option I.) is equivalent to R-FDM. We also remark that there is a need to modify R-FDM so that the minibatch stochastic coordinate descent method can be analyzed even when $X \neq \R^n$.
However, first we describe SCDM and make the following assumption in order to establish the equivalence of SCDM with $X=\R^n$ and R-FDM.

\begin{algorithm}[h!]
\caption{Stochastic  Coordinate Descent  Method (SCDM)}
\label{alg:sdca}
\begin{algorithmic}[1]
\STATE {\bf Input:} $f(x)$, $\{\vi{\omega}{k}\}_{k=0}^\infty$, diagonal matrix $W\succ 0$, $\vi{x}{0}$,
size of minibatch $\tau \in\{1,2,\dots, n\}$
\STATE {\bf Input:} $X = X_1 \times \dots \times  X_n$, where $X_i = [a,b]$ with
$ -\infty \leq a < b \leq +\infty $
\WHILE {$k\geq 0$:}
	\STATE
	choose $i \in \{1,2,\dots, n\}$ uniformly at random
   \STATE
    set $\vi{x}{k+1} = \vi{x}{k}$
 \STATE
 Option I:\\

    $\vc{\vi{x}{k+1}}{i}
 =    \arg\min_{\vc{x}{i} \in X_i}
        f(
        (
        \vc{\vi{x}{k}}{1},
        \vc{\vi{x}{k}}{2},
        \dots,
        \vc{\vi{x}{k}}{i-1},
        \vc{x}{i},
        \vc{\vi{x}{k}}{i+1},
        \dots,
        \vc{\vi{x}{k}}{n}
          )^T)	
  $	
   \label{algLineUpdate}
    \STATE
 Option II:\\
	 $\vi{x}{k+1}
 =   \proj{X}{\vi{x}{k} - \omega_k W^{-1} \nabla_i f(\vi{x}{k}) e_i}
 $
\ENDWHILE
\end{algorithmic}
\end{algorithm}

\begin{assumption}\label{asm:CSC}
The function $f$ is coordinate-wise strongly convex with respect to the norm $\|\cdot\|_W$ with parameter $\gamma>0$, if, for any $x\in X$ and any $i\in \{1,2,\dots,n\}$ we have
\begin{equation}\label{eq:CSC}
f(\vc{x}{1},\dots,
\vc{x}{i-1},\xi,\vc{x}{i+1},\dots,
\vc{x}{n})
 - f(x) + \nabla_i f(x) (\vc{x}{i}-\xi)
  \geq \gamma w_i |\xi-\vc{x}{i}|^2.
\end{equation}
\end{assumption}
Note that Assumption~\ref{asm:CSC} \emph{does not imply strong convexity of the function $f$}. For example, \eqref{eq:CSC} is satisfied for the Lasso problem or for the SVM dual problem whenever $\forall i: \|a_i\|>0$, and neither of those problems is strongly convex.

\begin{theorem}
\label{thm:sdca_is_RFDM}
Let Assumptions \ref{asm:wsc}, \ref{asm:lc} and
\ref{asm:CSC} hold.
If $X = \R^n$ then the Stochastic Coordinate Descent Method (SCDM)
(Algorithm \ref{alg:sdca} with Option I.) is equivalent to R-FDM with the parameters
$\beta^2 = 2 [(L_f^W)^2+1]
  +(n-1) r^2$, $\zeta = \gamma $ and $\omega_k = 1$,
where
$ r^2 = \max_i \frac{L_i^2}{w_i^2}$.
\end{theorem}
The following remark compares the result of the above theorem with the cyclic rule.
\begin{remark}
It was shown in \cite{luo1993error}
that for the cyclic coordinate descent method
(which is not randomized and hence
\eqref{eq:RFDM-1}-\eqref{eq:RFDM-3} hold deterministically) we have
$\omega_k^{\mbox{cyclic}}=1$, $\zeta^{\mbox{cyclic}} = \gamma$
and $(\beta^{\mbox{cyclic}})^2
 = (1+\sqrt{n} L_f^W)^2 = 1 + 2 \sqrt{n} L_f^W +
 n (L_f^W)^2$.
 For simplicity, let us assume that
 $W = \diag(L_1,L_2,\dots,L_n)$. Then
 $r^2=1$ and $L_f^W \in [1,n]$.
 For the cyclic coordinate descent method and SCDM, $\omega_k$ and $\zeta$ are the same.
 However, if we consider the worst case (when $L_f^W = n$)
 we have that
 $\beta^2 \sim \mathcal{O}(n^2)$, whereas
 $(\beta^{\mbox{cyclic}})^2 \sim \mathcal{O}(n^3)$.
 Also note that one iteration of
 cyclic coordinate descent requires $n$ coordinate updates, whereas SCDM updates just {\bf one} coordinate, and therefore each iteration of SCDM is $n$ times cheaper.
 In the other extreme, when $L_f^W = 1$ we have that
 both
 $\beta^2 \sim (\beta^{\mbox{cyclic}})^2
 \sim \mathcal{O}(n)$, but again we recall that one iteration of SCDM is $n$ times cheaper.
\end{remark}

It turns out that if $X \neq \R^n$ then SCDM does not fit the R-FDM framework because $\nabla_i f(x_k)$ cannot be bounded by $\|\vi{x}{k}-\vi{x}{k+1}\|_W$. Thus, there is a need to modify R-FDM such that the SCDM algorithm can be analyzed for bounded problems.

The natural modification to R-FDM, which would allow SCDM to fit the R-FDM framework is the following: at each iteration $k$ we require that in \eqref{eq:RFDM-1}, only a subset of coordinates of the vector $\vi{x}{k}$ are updated. This can be achieved by the following method.
\begin{definition}
[Randomized Coordinate Feasible Descent Method (RC-FDM)]
Let $X = X_1 \times  \dots \times X_n$,
where $X_i$ are intervals.
A sequence $\{\vi{x}{k}\}_{k=0}^{\infty}$
is
generated by RC-FDM if
there exists $\beta\geq0$, $\zeta > 0$
and $\{\vi{\omega}{k}\}_{k=0}^\infty$
with $\min_{k} \vi{\omega}{k} \geq  \bar \omega > 0$
such that
 for every iteration $k$, the following are satisfied
\begin{align}
\label{eq:RCFDM-1}
\vi{x}{k+1}
 &= \proj{X}{
 \vi{x}{k} - \vi{\omega}{k}
  W^{-1}
  \vsubset{(\nabla f(\vi{x}{k})
   - \vi{z}{k})}{i}
  },\\
  \label{eq:RCFDM-2}
  (\|
 \vsubset{(
  \vi{z}{k})}{i}
  \|_W^*)^2
  &\leq \beta^2  \|\vi{x}{k}
    -\vi{x}{k+1}\|_W^2,
    \\
    \label{eq:RCFDM-3} f(\vi{x}{k+1})
 &\leq f(\vi{x}{k})
  - \zeta   \|\vi{x}{k} - \vi{x}{k+1}\|_W^2,
\end{align}
where $i$ is a coordinate
selected uniformly at random
from the set $\{1,2,\dots,n\}$, $\vsubset{x}{i}$ is a vector
whose elements $j\neq i$ are set to $0$
and $z_k$ is some fixed vector at iteration $k$.
\end{definition}
Now, we can show that even if $X \neq \R^n$, SCDM is RC-FDM. The first theorem holds if Option I. is  used in Algorithm \ref{alg:sdca} and the second theorem holds if Option II. is used.
\begin{theorem}
\label{thm:sdca_is_RCFDM}
Let Assumptions \ref{asm:wsc}, \ref{asm:lc}  and
\ref{asm:CSC} hold.
If $X = X_1 \times  \dots \times X_n$,
where $X_i$ are intervals then the
Stochastic Coordinate Descent Method
in Algorithm \ref{alg:sdca} with Option I.
is RC-FDM with
$\beta^2 =  2 [(L_f^W)^2+1]$, $\zeta =\gamma$, and $\omega_k = 1 $.
\end{theorem}
\begin{theorem}
\label{thm:sdca_is_RCFDM2}
Let Assumptions \ref{asm:wsc}, \ref{asm:lc} and
\ref{asm:CSC} hold.
If $X = X_1 \times  \dots \times X_n$,
where $X_i$ are intervals then the
Stochastic Coordinate Descent Method
in Algorithm \ref{alg:sdca} with Option II.
is RC-FDM with
$\vi{z}{k} = 0$, $\zeta = \gamma$, $\beta=0$, $\omega_k = 1$, and $W=\diag(L_1,L_2,\dots,L_n)$.
\end{theorem}

\section{Convergence Analysis}
\label{sec:CA}
In \cite{necoara2015linear}
they proved linear convergence for FDM
under Assumptions \ref{asm:wsc}
 and \ref{asm:lc}.
The following theorem shows that a linear convergence rate can also be established for R-FDM.
\begin{theorem}[Linear Convergence of R-FDM]
\label{thm:LC:R-FDM}
Let Assumptions \ref{asm:wsc} and \ref{asm:lc} hold.
If the sequence $\{\vi{x}{k}\}_{k=0}^\infty$ is produced by
R-FDM (i.e. \eqref{eq:RFDM-1}-\eqref{eq:RFDM-3} are satisfied)
then
 \begin{align*}
 \Exp[f(\vi{x}{k}) - f^*]
&\leq
 \left(\frac{c}{1+c}\right)^k  \left(  f(\vi{x}{0})  -f^*  \right),\tagthis \label{afw34r32efae}
\end{align*}
where
\begin{equation}
c = \frac2{\kappa_f \zeta }
\left(
(L_f^W+\tfrac1{\bar  \omega })^2
+
\beta^2 \right).
\end{equation}
\end{theorem}

The next theorem establishes a linear convergence rate for
RC-FDM.

\begin{theorem}[Linear Convergence of RC-FDM]
\label{thm:LC:RC-FDM}
Let $X = X_1 \times  \dots \times X_n$,
where $X_i$ are intervals.
Further, let Assumptions \ref{asm:wsc} and \ref{asm:lc} hold.
Let the sequence $\{\vi{x}{k}\}_{k=0}^\infty$ be produced by
RC-FDM (i.e. \eqref{eq:RCFDM-1}-\eqref{eq:RCFDM-3} are satisfied),
then for $z_k \neq 0$ there exists $c\in (0,1)$ such that
all $k$
 \begin{align*}\label{eq:CT-RC-FDMasdfadfa}\tagthis
\Exp[f(\vi{x}{k})-f^*]
\leq
 (1-c)^k \left(f(\vi{x}{0})-f^*
\right).
\end{align*}
Moreover, if for all $k$ we have $\vi{z}{k} \equiv 0$, and $\frac1{\omega_k} \geq
\max_{i} \frac{L_i}{w_i}$, then
$c = \frac{2 \bar \omega \kappa}{n(2 \bar \omega \kappa+1)}$ with
 \begin{align*}\label{eq:CT-RC-FDM}\tagthis
\Exp[f(\vi{x}{k})-f^*]
\leq
 (1-c)^k \left(f(\vi{x}{0})-f^*
 +\frac1{2\bar \omega} \|\vi{x}{0} - \vi{\bar x}{0}\|_W^2\right).
\end{align*}
\end{theorem}
\subsection{Comparison with the Results in Related Literature}
In Theorem~\ref{thm:LC:RC-FDM} we established the linear convergence of RC-FDM for any $z_k$. We will now compare our result with the one presented in \cite{liu2014asynchronous} for the projected coordinate gradient descent
algorithm. Note that the projected coordinate gradient descent
algorithm fits the RC-FDM framework exactly. We also note that the result in 
\cite{liu2014asynchronous} only holds for $z_k = 0$, so our result is more general.
Further, even though the paper \cite{liu2014asynchronous}
considers an asynchronous implementation,
where the update computed at iteration $k$ is based on gradient information at a point up to $\tau$ iterations old, if $\tau=0$ then their method fits into the RC-FDM framework.
One of the benefits of our work is that more general norms can be used. So, for simplicity, and to match with the work in \cite{liu2014asynchronous}, let us assume that $L_i = 1$ for all $i$ and we also choose $w_i = 1$ for all $i$.
(This is the case e.g. for the SVM dual problem).
The geometric rate in \eqref{eq:CT-RC-FDM}
in our work is then
$\displaystyle
 1 - \tfrac{  \kappa}{n ( \kappa +\frac12)}
$
and from Theorem 4.1 in \cite{liu2014asynchronous}
for $\tau=0$ we obtain that the geometric rate is
$\displaystyle
 1 - \tfrac{ \kappa}{n (\kappa +L_{\max})},
$
where $L_{\max}\geq 1$ is such that
$$
\|\nabla f(x) - \nabla f(x+\delta e_i)\|_\infty \leq L_{\max} |\delta|
$$
holds $\forall x \in \R^n$, $\delta \in \R$ and $i\in \{1,2,\dots,n\}$.
Hence, in this case our convergence results are better.

In \cite{necoara2015linear} the author provided a linear convergence rate for deterministic FDM. It is shown in Theorem 3.2 in \cite{necoara2015linear} that
the coefficient of the linear rate is
$\displaystyle
1-\tfrac{   \zeta }{\zeta+\rho }
$
where $\rho = \frac1{\kappa_f} (L_f + \frac1{\bar \omega}+\beta)^2$
whereas, in Theorem \ref{thm:LC:RC-FDM} of this work,
from \eqref{afw34r32efae}
we see that the coefficient is
the same but with a different $\rho$. To be precise, in our case we have
$
\bar \rho  = \frac2{\kappa_f   }
\left(
(L_f^W+\tfrac1{\bar  \omega })^2
+
\beta^2 \right).
$
Our result can be better or worse than that in \cite{necoara2015linear}, depending
on the values of $L_f^W, \bar \omega$ and $\beta$, but our results holds for R-FDM, which is broader than FDM.

\section{Global Error Bound Property}

\label{sec:GEB}

In this Section we describe a class of problems that satisfies the Global Error Bound (GEB) property.
We show that this implies the weak strong convexity property and we compare the convergence rate obtained in this paper with several results in the current literature derived for problems obeying the GEB. We begin by defining the projected gradient.
\begin{definition}[Projected Gradient]
For any $x\in \R^n$ let us define the projected gradient as follows:
	\begin{equation}
		\n^+ f(x) := x-\proj{\mathcal X}{x-\n f(x)}.
	\end{equation}
\end{definition}
Note that projected gradient is zero at $x$ if and only if $x$ is an optimal solution of \eqref{eq:P}. Also, we will employ the projected gradient to define an error bound, which measures the distance between $x$ and the optimal solution.
Now, we are ready to define a {\em global error bound} as follows.
\begin{definition}[Definition 6 in \cite{wang2014iteration}]
An optimization problem admits a global error bound if there is a constant $\eta$ such that
\begin{equation}\label{Eq:GEBDef}
\|x - \bar x\| \leq \eta_f  \|\n^+ f(x)\|_W^*,
\quad \forall x\in X.
\end{equation}
A relaxed condition called {\em the global error bound from the beginning} is if the above inequality holds only for $x \in X$ such that $f(x)-f(\bar x) \leq M$, where $M$ is a  constant, and usually we have that $M = f(\vi{x}{0}) - f^*$.

\end{definition}

Let us consider a special instance of \eqref{eq:P} when $X$ is polyhedral set, i.e.
\begin{equation}\label{Eavfrwaevfravfa}
X = \{x\in \R^n:  B x \leq c\},
\end{equation}
and the function $f$ has the following structure
\begin{equation}\label{Eq:afoejrgoeajnwvfa}
f(x) = h(Ax) + q^Tx,
\end{equation}
where $B \in \R^{l \times n}$,
 $A\in \R^{d \times n}$, $h$ is a $\sigma_h$ strongly convex function
and $f$ satisfies Assumption \ref{asm:lc}.
We also assume that there exists an optimal solution
and hence the optimal solution set $X^*$ is assumed to be non-empty \cite{wang2014iteration}.
It is easy to observe that if $f$ is strongly convex, then
\eqref{eq:WSC} is trivially satisfied.
Just recently, \cite{necoara2015linear}
showed that
if  \eqref{Eq:GEBDef}
is satisfied,
then
\eqref{eq:WSC}
is satisfied with
\begin{equation}\label{afowjeovkejlrvnrea}
\kappa_f = \frac{L_f^W}{2 \eta_f^2}.
\end{equation}
For problem
\eqref{Eq:afoejrgoeajnwvfa}
it was discussed in
\cite{wang2014iteration}
that
\begin{equation}
\label{avwefwfvafvaew}
\eta_f
=\theta^2 (1+L_f^W)\left(
\frac{1+2\|\nabla h(A \bar x)\|^2}{\sigma_h}+4M\right)
+2 \theta \|\nabla f(\bar x)\|,
\end{equation}
where
$\theta$ is a constant from the Hoffman bound
\cite{hoffman1952approximate,li1993sharp,robinson1973bounds}
defined as follows
\begin{equation}\label{eqavfefreafvreav}
\theta \eqdef
\sup_{u,v}\left\{
\left\|\begin{pmatrix}u\\v\end{pmatrix}\right\|
 \left|
 \begin{array}{l}
 \left\|B^T u + \begin{pmatrix}
     A\\q^T
 \end{pmatrix}^T v\right\| = 1, u\geq 0\\
 \mbox{and the corresponding rows of }B, A  \mbox{ to }
 u,v\mbox{'s}\\
 \mbox{non-zero elements are linearly independent.}
\end{array}
   \right.
 \right\}.
\end{equation}
Note that the constant $\theta$ can be very big (we will provide a brief discussion on this in Section \ref{sec:SDCA}).

In \cite{necoara2015linear} they derived that for problem
\eqref{Eq:afoejrgoeajnwvfa},
the weak strong convexity property \eqref{eq:WSC}
holds with
\begin{equation}
\label{Eq:afjoiewjfoafew}
\kappa_f
= \frac{\sigma_h}{2 \theta^2}.
\end{equation}
Note that  $\kappa_f$ given in
\eqref{Eq:afjoiewjfoafew}
is $\mathcal{O}(\theta^2)$ whereas
$\kappa_f$ obtained from
\eqref{afowjeovkejlrvnrea}
is of the order $\theta^4$. Therefore we will compare our results using the latter estimates of $\kappa_f$.

\subsection{Comparison with the Results in Related Literature}

In Theorem 8 in \cite{wang2014iteration}, under the global error bound property, it is proven that FDM converges at a linear rate:
$f(\vi{x}{k+1})-f^* \leq (1-\frac{1}{\bar c+1})
 (f(\vi{x}{k})-f^*)$,
 with\footnote{In \cite{wang2014iteration}
 it was shown that  \eqref{avwefwfvafvaew},
 in some special cases (e.g. when $X=\R^n$), is
 $\eta_f = \theta^2\frac{1+L_f^W}{\sigma_h}$.}
\begin{align*}
c
 &=\frac1{\zeta} (L_f^W + \frac1{\bar \omega} + \beta)
   (1+ \eta_f (\frac1{\bar \omega}+ \beta))
=
\frac1{\zeta} (L_f^W + \frac1{\bar \omega} + \beta)
   (1+  \theta^2\frac{1+L_f^W}{\sigma_h} (\frac1{\bar \omega}+ \beta))
\\
&\sim \mathcal{O}\left(
\frac{\theta^2}{\zeta \sigma_h}
(1+L_f^W)(
\frac1{\bar \omega}+\beta)(L_f^W + \frac1{\bar \omega} + \beta)
\right).
\end{align*}
   From Theorem \ref{thm:LC:R-FDM} in this work, we have linear convergence of RC-FDM with the coefficient
\begin{align*}
c = \frac2{\kappa_f \zeta }
\left(
(L_f^W+\tfrac1{\bar  \omega })^2
+
\beta^2 \right)
\overset{\eqref{Eq:afjoiewjfoafew}}{=}
\frac{4 \theta^2}{\sigma_h\zeta }
\left(
(L_f^W+\tfrac1{\bar  \omega })^2
+
\beta^2 \right).
\end{align*}
These coefficients are very similar, but FDM \cite{wang2014iteration} covers only cyclic coordinate descent and not a randomized coordinate descent method (which is covered by Theorem \ref{thm:LC:R-FDM}).

\section{Linear Convergence Rate of SDCA for Dual of SVM}
\label{sec:SDCA}

In this Section we show that the SDCA algorithm (which is SCDM applied to \eqref{eq:tempdual}) achieves a linear convergence rate for the duality gap.
This improves upon the result obtained in
\cite{shalev2013stochastic,PrimalDualParalle,TRN:minibatchICML}
where only a sublinear rate was derived.

Let us assume, for simplicity, that
in problem \eqref{eq:prime}
for all $i\in \{1,2,\dots,n\}$ it holds that
$\|a_i\|\leq 1$.
Then from  \cite{PrimalDualParalle,TRN:minibatchICML}
we have that for any $x \in \R^n, s\in [0,1]$
and the function $f$ defined in \eqref{eq:tempdual}
we have
\begin{equation}\label{eqafcewfvaevw}
 f(x)-f^* \geq s G(x) -s^2   \frac{ \sigma^2}{2\lambda},
\end{equation}
where $f^*$ denotes the optimal value of \eqref{eq:tempdual},
$A=[a_1, a_2, \dots, a_n], \sigma^2 = \frac1n\| X \| \in [\frac1n,1]$
and $G(x)$ is the duality gap at the point $x$, which is defined as
$G(x) \eqdef P(\frac1{\lambda n} Ax)+ f(x).$

Let us remark that SDCA for problem \eqref{eq:tempdual}
is equivalent to RC-FDM, where the constants in \eqref{eq:RCFDM-1}-\eqref{eq:RCFDM-3}
are given as follows:
$\vi{z}{k} = 0$, $\beta^2=0$,
$w_i = L_i  = \frac1{\lambda n^2} \|a_i\|^2$, and
$\omega_k=1$.
Hence, if we choose $\vi{x}{0} = {\bf 0}$ then from Theorem
\ref{thm:LC:RC-FDM}
we have
that
$\Exp[f(\vi{x}{k})-f^*]
\leq (1-c)^k
\left(
f({\bf 0}) - f^* +  \|x^*\|_L^2\right)
$
with
$
c = \frac{2  \kappa_f}{n(2  \kappa_f+1)}
$.

Now, we see that rearranging \eqref{eqafcewfvaevw} gives
\begin{equation}\label{afweverwfav}
G(x) \overset{\eqref{eqafcewfvaevw}}{\leq} s   \frac{ \sigma^2}{2\lambda} + \frac1s (f(x)-f^*).
\end{equation}
 If we want to achieve
 $G(x) \leq \epsilon$
 it is sufficient to choose
 both terms on right hand side of
 \eqref{afweverwfav}
 to be $\leq \frac{\epsilon}{2}$.
 Hence, we can set
 $s =\min\{1,  \frac{\epsilon\lambda}{ \sigma^2} \}$.
 All we have to do now is to choose
 $k$ such that
 $ f(\vi{x}{k})-f^*  \leq s \frac{\epsilon}{2}$.
In the following theorem we establish linear convergence of the duality gap $G(x)$
for the SDCA algorithm.
\begin{theorem}
Let $s =\min\{1,  \frac1{\epsilon\lambda}{\sigma^2} \}$
and let $K$ be such that
$$K
 \geq
n \left(1+  \frac{ 1 }{2  \kappa_f}\right)
 \log
\frac {2\left(
f({\bf 0}) - f^* +  \|x^*\|_L^2\right)}
 { s \epsilon }.  $$
 Then if the SDCA algorithm is applied to problem
 \eqref{eq:tempdual} to produce $\{\vi{x}{k}\}_{k=0}^\infty$, then $\forall k\geq K$ we have
 that
 $\Exp[G(\vi{x}{k})] \leq \epsilon$.
\end{theorem}

Let us now comment on the size of the parameter $\kappa_f \overset{ \eqref{Eq:afjoiewjfoafew}} {=} \frac{\sigma_h}{2 \theta^2}$.
  In our case, $X$ is the polyhedral set \eqref{Eavfrwaevfravfa} defined by
  $B=\begin{pmatrix}
  -I_n & I_n
  \end{pmatrix}^T,$ and
    $c=({\bf 0}^T, {\bf 1}^T)^T$, where
    $I_n \in \R^{n \times n}$ is the identity matrix.  Because of this structure \eqref{eqavfefreafvreav} simplifies to
\begin{equation}\label{aesvevgfeavgsa}
\theta \eqdef
\sup_{u,n}\left\{
\left\|\begin{pmatrix}u\\v\end{pmatrix}\right\|
 \left|
 \begin{array}{l}
 \left\| I_n  u + \begin{pmatrix}
     A\\q^T
 \end{pmatrix}^T v\right\| = 1\\
 \mbox{and the corresponding rows of } I_n, A  \mbox{ to }
 u,v\mbox{'s}\\
 \mbox{non-zero elements are linearly independent.}
\end{array}
   \right.
 \right\}.
\end{equation}
To show that $\theta$ can be very large, let us assume that
two rows of the matrix $A$ are highly correlated (in this case rows corresponds to features).
We denote these two rows by $A_1$ and $A_2$, and let us assume that $A_1 = A_2 + \delta e_1$.
Then we can chose
$v = (-\frac1\delta, \frac1\delta, 0, \dots,0)^T$ and $u={\bf 0}$. This particular choice is feasible in optimization problem
\eqref{aesvevgfeavgsa}
and hence is imposing  a lower-bound on $\theta$:
$\theta \geq \frac{\sqrt{2}}{|\delta|}$.

 \section{Summary}\label{S_Summary}
In this paper we have extended the framework of the feasible descent method FDM into a randomized and a randomized coordinate FDM framework. We have provided a linear convergence rate (under the weak strong convexity assumption) for both methods and we have shown that the convergence rates are similar to the deterministic/non-randomized FDM. We showed that for the cyclic coordinate descent method the coefficients in FDM are worse or similar to the stochastic coordinate descent method (and hence the theory tells us that they converge at roughly the same speed), but each iteration of the stochastic coordinate descent method is $n$-times cheaper. We concluded the paper with a result showing that, for the SDCA algorithm applied to the dual of the linear SVM, the duality gap converges linearly.

{

\bibliographystyle{plain}
\bibliography{citations}
}

\clearpage
\appendix
\section{Proofs}


\subsection{Proof of Theorem \ref{thm:sdca_is_RFDM}}

Let us define an auxiliary vector $\tilde x$ such that
\begin{equation}
\label{eqasdfasfdsafa}
\vc{\tilde x}{i}
 =  \arg\min_{\vc{x}{i} \in X_i}
        f(
        (
        \vc{\vi{x}{k}}{1},
        \vc{\vi{x}{k}}{2},
        \dots,
        \vc{\vi{x}{k}}{i-1},
        \vc{x}{i},
        \vc{\vi{x}{k}}{i+1},
        \dots,
        \vc{\vi{x}{k}}{n}
          )^T).	
\end{equation}
Then we can see
that if  coordinate $i$ is chosen during iteration $k$ in Algorithm \ref{alg:sdca}
then
\begin{equation}
\vc{\vi{x}{k+1}}{j}
 = \begin{cases}
  \vc{\vi{x}{k}}{j},
  &\mbox{if }\ j\neq i,\\
 \vc{\tilde x}{i},&\mbox{otherwise.}
 \end{cases}\label{asfasfavgaa}
\end{equation}
If coordinate $i$ is chosen during iteration $k$,
then the optimality conditions for Step \ref{algLineUpdate} of Algorithm
\ref{alg:sdca}, give us that
\begin{equation}\label{Alg1optconds}
  \vc{\vi{x}{k+1}}{i}
   = \proj{X_i}{
   \vc{\vi{x}{k+1}}{i}
   - \frac1{w_i} \nabla_i f ( \vi{x}{k+1})
     }.
\end{equation}
Moreover, by \eqref{asfasfavgaa}, for $j\neq i$
we have that
$\vc{\vi{x}{k}}{j}=\vc{\vi{x}{k+1}}{j}$
which is possible only if $\vc{\vi{z}{k}}{j} = \nabla _j f(\vi{x}{k})$.

Note that $\vi{x}{k+1}$ is a random variable, which depends on $i$ and $\vi{x}{k}$ only.
Therefore, we can define a random
$\vi{z}{k}$ such that the $i$-th coordinate is
\begin{equation}\label{Proofzki}
 \vc{\vi{z}{k}}{i}
  = \nabla_i f(\vi{x}{k}) - \nabla_i f((
        \vc{\vi{x}{k}}{1},
        \vc{\vi{x}{k}}{2},
        \dots,
        \vc{\vi{x}{k}}{i-1},
        \vc{\tilde x}{i},
        \vc{\vi{x}{k}}{i+1},
        \dots,
        \vc{\vi{x}{k}}{n}
          )^T)
   + w_i ( \vc{\vi{x}{k}}{i}
     - \vc{\tilde x }{i})
\end{equation}
and the $j$-th coordinate (for $j\neq i$) is defined as
$\vc{\vi{z}{k}}{j} =
\nabla _j f(\vi{x}{k})$.
It is easy to verify that
for $\vi{z}{k}$ defined above,
condition \eqref{eq:RFDM-1} holds.
Now, we will compute
$\Exp[
(\|\vi{z}{k} \|_W^*)^2]
$.
We have that if the $i$-th coordinate is chosen then
\begin{align*}
\frac1{w_i}(\vc{\vi{z}{k}}{i})^2
&=\frac1{w_i}\Big(
\nabla_i f(\vi{x}{k}) - \nabla_i f((
        \vc{\vi{x}{k}}{1},
        \vc{\vi{x}{k}}{2},
        \dots,
        \vc{\vi{x}{k}}{i-1},
        \vc{\tilde x}{i},
        \vc{\vi{x}{k}}{i+1},
        \dots,
        \vc{\vi{x}{k}}{n}
          )^T)
   + w_i ( \vc{\vi{x}{k}}{i}
     - \vc{\tilde x }{i})
    \Big)^2
\\
&\leq
\frac2{w_i}\Big(
\nabla_i f(\vi{x}{k}) - \nabla_i f((
        \vc{\vi{x}{k}}{1},
        \vc{\vi{x}{k}}{2},
        \dots,
        \vc{\vi{x}{k}}{i-1},
        \vc{\tilde x}{i},
        \vc{\vi{x}{k}}{i+1},
        \dots,
        \vc{\vi{x}{k}}{n}
          )^T)\Big)^2
   + 2 w_i  ( \vc{\vi{x}{k}}{i}
     - \vc{\tilde x }{i}
    )^2
\\
&\leq
2
\big(\|
\nabla  f(\vi{x}{k}) - \nabla  f((
        \vc{\vi{x}{k}}{1},
        \vc{\vi{x}{k}}{2},
        \dots,
        \vc{\vi{x}{k}}{i-1},
        \vc{\tilde x}{i},
        \vc{\vi{x}{k}}{i+1},
        \dots,
        \vc{\vi{x}{k}}{n}
          )^T) \|_W^*\big)^2
   + 2 w_i  ( \vc{\vi{x}{k}}{i}
     - \vc{\tilde x }{i}
    )^2
\\
&\overset{\eqref{eq:globalLipConstant}}
{\leq}
2
(L_f^W \|
  \vi{x}{k}  -  (
        \vc{\vi{x}{k}}{1},
        \vc{\vi{x}{k}}{2},
        \dots,
        \vc{\vi{x}{k}}{i-1},
        \vc{\tilde x}{i},
        \vc{\vi{x}{k}}{i+1},
        \dots,
        \vc{\vi{x}{k}}{n}
          )^T \|_W )^2
   + 2 w_i  ( \vc{\vi{x}{k}}{i}
     - \vc{\tilde x }{i}
    )^2
\\
&=
 2 (L_f^W)^2 w_i
(  \vc{\vi{x}{k}}{i}
     - \vc{\tilde x }{i} )^2
   + 2 w_i  ( \vc{\vi{x}{k}}{i}
     - \vc{\tilde x }{i}
    )^2
=  2 [(L_f^W)^2+1] w_i
(  \vc{\vi{x}{k}}{i}
     - \vc{\tilde x }{i} )^2,
     \tagthis\label{asfddaasdf232dfdsafsafa}
\end{align*}
otherwise
\begin{align*}
\frac1{w_i} (\vc{\vi{z}{k}}{i})^2
&= \frac1{w_i} (\nabla_i f(\vi{x}{k}))^2.
\end{align*}
Hence,
we obtain that
\begin{align*}
\Exp[ (\|\vi{z}{k} \|_W^*)^2 ]
&\overset{\eqref{asfddaasdf232dfdsafsafa}}{\leq}
 \sum_{i=1}^n
  \frac1n 2 [(L_f^W)^2+1] w_i
(  \vc{\vi{x}{k}}{i}
     - \vc{\tilde x }{i} )^2
   +\frac{n-1}n \sum_{i=1}^n \frac1{w_i} (\nabla_i f(\vi{x}{k}))^2.
   \tagthis \label{af3w43r24r331}
\end{align*}

From the optimality condition of Step \ref{algLineUpdate} of Algorithm \ref{alg:sdca}, and the fact that $X_i = \R$, we know that for all $i$ the following holds:
\begin{equation}
\nabla_i f(\vc{\vi{x}{k}}{1},\dots,
\vc{\vi{x}{k}}{i-1},\vc{\tilde x}{i},
\vc{\vi{x}{k}}{i+1},\dots,\vc{\vi{x}{k}}{n}) = 0.
\end{equation}
Therefore
$\forall i$ we have
\begin{align*}
\frac1{w_i} (\nabla_i f(\vi{x}{k}))^2
&=
\frac1{w_i} (\nabla_i f(\vi{x}{k})-
\nabla_i f(\vc{\vi{x}{k}}{1},\dots,
\vc{\vi{x}{k}}{i-1},\vc{\tilde x}{i},
\vc{\vi{x}{k}}{i+1},\dots,\vc{\vi{x}{k}}{n})
)^2
\\
&\overset{\eqref{eq:CWLC}}{\leq}
\frac1{w_i} L_i^2(\vc{\tilde x}{i} - \vc{ \vi{x}{k}}{i})^2
= \frac1{w_i^2} L_i^2
w_i (\vc{\tilde x}{i} - \vc{ \vi{x}{k}}{i})^2.
\end{align*}
If we denote by
$ r^2 = \max_i \frac{L_i^2}{w_i^2}$, then we obtain
from \eqref{af3w43r24r331}
\begin{align*}
\Exp[ (\|\vi{z}{k} \|_W^*)^2 ]
&\overset{\eqref{asfddaasdf232dfdsafsafa}}{\leq}
 \sum_{i=1}^n
\left(
  \tfrac1n 2 [(L_f^W)^2+1]
  +\tfrac{n-1}n r^2
  \right)
   w_i
(  \vc{\vi{x}{k}}{i}
     - \vc{\tilde x }{i} )^2
\\
&=
\left(
  \tfrac1n 2 [(L_f^W)^2+1]
  +\tfrac{n-1}n r^2
  \right)
 \sum_{i=1}^n
   w_i
(  \vc{\vi{x}{k}}{i}
     - \vc{\tilde x }{i} )^2
     \\
&=
\left(
  2 [(L_f^W)^2+1]
  +(n-1) r^2
  \right)
   \tfrac1n
 \sum_{i=1}^n
   w_i
(  \vc{\vi{x}{k}}{i}
     - \vc{\tilde x }{i} )^2
          \\
&=
\left(
  2 [(L_f^W)^2+1]
  +(n-1) r^2
  \right)
 \Exp[ \|\vi{x}{k} - \vi{x}{k+1}\|_W^2]
\end{align*}
and  we can conclude that
\eqref{eq:RFDM-2}
holds with $\beta^2 = 2 [(L_f^W)^2+1]
  +(n-1) r^2$.

Now, it remains to show \eqref{eq:RCFDM-3}.
From  \eqref{eqasdfasfdsafa} we know that
\begin{equation}
\nabla_if(
        (
        \vc{\vi{x}{k}}{1},
        \vc{\vi{x}{k}}{2},
        \dots,
        \vc{\vi{x}{k}}{i-1},
        \vc{\tilde x}{i},
        \vc{\vi{x}{k}}{i+1},
        \dots,
        \vc{\vi{x}{k}}{n},
          )^T)
          (\vc{\tilde x}{i} - \vc{\vi{x}{k}}{i}) \leq 0.
          \label{Asdfwvfewavgfearvs}
\end{equation}

Therefore from
 \eqref{eq:CSC}
 with $\xi = \vc{\vi{x}{k}}{i}$
 and
 $x = (
 \vc{\vi{x}{k}}{1},
 \dots,
 \vc{\vi{x}{k}}{i-1},
   \vc{\tilde x}{i},
   \vc{\vi{x}{k}}{i+1},
   \dots,\vc{\vi{x}{k}}{n})^T
   \overset{\eqref{asfasfavgaa}}{=} \vi{x}{k+1} $
we have that
\begin{align*}
 f(\vi{x}{k})
  - f(\vi{x}{k+1})
   \geq \gamma w_i
   |
   \vc{  \vi{x}{k} }{i} -
   \vc{ \vi{x}{k+1} }{i}  |^2
   +
   \nabla_i f(\vi{x}{k+1})
   (\vc{  \vi{x}{k} }{i} -
   \vc{ \vi{x}{k+1} }{i})
  \overset{\eqref{Asdfwvfewavgfearvs}}{\geq}
  \gamma w_i
   |
   \vc{  \vi{x}{k} }{i} -
   \vc{ \vi{x}{k+1} }{i}  |^2.
   \label{eq:1Afewvfavavafdsvsd}
   \tagthis
\end{align*}
Therefore
\begin{align*}
  f(\vi{x}{k})
  - f(\vi{x}{k+1})
  &\overset{\eqref{eq:1Afewvfavavafdsvsd}}{\geq} \gamma
    w_i
   |
   \vc{  \vi{x}{k} }{i} -
   \vc{ \vi{x}{k+1} }{i}  |^2
   = \gamma  \|\vi{x}{k} - \vi{x}{k+1}\|_W^2.
\end{align*}
 and by taking expectation on both sides of the above,
 \eqref{eq:RFDM-3} follows with $\zeta = \gamma$.


\subsection{Proof of Theorem \ref{thm:sdca_is_RCFDM}}

This proof is very similar to the proof of Theorem \ref{thm:sdca_is_RFDM}.
Let us define an auxiliary vector $\tilde x$ in the same way as in \eqref{eqasdfasfdsafa}.
Then we can see that if coordinate $i$ is chosen during iteration $k$ in Algorithm \ref{alg:sdca} then \eqref{asfasfavgaa} holds, and the optimality conditions for
Step \ref{algLineUpdate} of Algorithm \ref{alg:sdca} imply that \eqref{Alg1optconds} holds.

Note that $\vi{x}{k+1}$ is a random variable which depends on $i$ and $\vi{x}{k}$ only.
Therefore, we can define $\vi{z}{k}$ such that $i$-th coordinate is given by \eqref{Proofzki}. It is easy to verify that for $\vi{z}{k}$ defined in \eqref{Proofzki}, the condition \eqref{eq:RCFDM-1} holds. Now, let us compute
$ (\| \vsubset{(\vi{z}{k})}{i} \|_W^*)^2 $. We have that
\begin{equation*}
(\|
 \vsubset{(
  \vi{z}{k})}{i}
  \|_W^*)^2 =
\frac1{w_i}(\vc{\vi{z}{k}}{i})^2
\overset{\eqref{asfddaasdf232dfdsafsafa}}{\leq}
2 [(L_f^W)^2+1] w_i
(  \vc{\vi{x}{k}}{i}
     - \vc{\tilde x }{i} )^2
     \overset{\eqref{asfasfavgaa}}{=}
2 [(L_f^W)^2+1] \|  \vc{\vi{x}{k}}{i}
     - \vc{ \vi{x}{k+1} }{i} \|_W^2.
\end{equation*}
Therefore, we conclude that \eqref{eq:RCFDM-2} holds with $\beta^2 = 2 [(L_f^W)^2+1]$.

Now, it remains to show \eqref{eq:RCFDM-3}. Again from \eqref{eqasdfasfdsafa} we know that \eqref{Asdfwvfewavgfearvs} holds. Therefore from \eqref{eq:CSC}
 with $\xi = \vc{\vi{x}{k}}{i}$
 and
 $x = (
 \vc{\vi{x}{k}}{1},
 \dots,
 \vc{\vi{x}{k}}{i-1},
   \vc{\tilde x}{i},
   \vc{\vi{x}{k}}{i+1},
   \dots,\vc{\vi{x}{k}}{n})^T
   \overset{\eqref{asfasfavgaa}}{=} \vi{x}{k+1} $
we have \eqref{eq:1Afewvfavavafdsvsd}. Therefore $f(\vi{x}{k})
  - f(\vi{x}{k+1})
  \overset{\eqref{eq:1Afewvfavavafdsvsd}}{\geq} \gamma
    w_i
   |
   \vc{  \vi{x}{k} }{i} -
   \vc{ \vi{x}{k+1} }{i}  |^2
   = \gamma  \|\vi{x}{k} - \vi{x}{k+1}\|_W^2$, so \eqref{eq:RCFDM-3} holds with $\zeta = \gamma$.

\subsection{Proof of Theorem \ref{thm:LC:R-FDM}}

This proof is based on the proof of Theorem 3.2 in \cite{necoara2015linear}.
We can write the optimality conditions
for $\vi{x}{k+1}$ from \eqref{eq:RFDM-1}
and using the definition of a projection given in
\eqref{Eq:pd}.
We have that $\forall x\in X$, the following inequality holds
\begin{equation}
\label{eq:asfewfawefvewa}
\ve{
W\left(\vi{x}{k+1}-
\vi{x}{k} + \vi{\omega}{k}
  W^{-1}
  (\nabla f(\vi{x}{k})
   - \vi{z}{k}) \right)}{ x - \vi{x}{k+1} }\geq 0.
\end{equation}
Now, using the convexity of $f$ we obtain that
\begin{align*}
 f(\vi{x}{k+1}) - f^*
&=
 f(\vi{x}{k+1}) - f(\vi{\bar x}{k+1})
\leq
 \ve{\nabla f(\vi{x}{k+1})}{ \vi{x}{k+1} - \vi{\bar x}{k+1}}
\\
&=
 \ve{
\nabla f(\vi{x}{k+1})
-
\nabla f(\vi{x}{k})
+
\nabla f(\vi{x}{k})
}{ \vi{x}{k+1} - \vi{\bar x}{k+1}}.
\tagthis \label{eqasfdasfa}
\end{align*}
Plugging $x=\vi{\bar x}{k+1}$ into \eqref{eq:asfewfawefvewa} we obtain
\begin{equation}
\label{eq:asfsafa}
\ve{
\tfrac1{\vi{\omega}{k}}
W (\vi{x}{k+1}-
 \vi{x}{k}) -
    \vi{z}{k}   }{ \vi{\bar x}{k+1} - \vi{x}{k+1} }
   \geq \ve{
   \nabla f(\vi{x}{k})
     }{   \vi{x}{k+1}-\vi{\bar x}{k+1} }.
\end{equation}
Plugging this into \eqref{eqasfdasfa}
gives us that
\begin{align*}
 f(\vi{x}{k+1}) - f(\vi{\bar x}{k+1})
&\overset{\eqref{eqasfdasfa},\eqref{eq:asfsafa}}{\leq}
 \ve{
\nabla f(\vi{x}{k+1})
-
\nabla f(\vi{x}{k})
-
\tfrac1{\vi{\omega}{k}}
W (\vi{x}{k+1}-
 \vi{x}{k}) +
    \vi{z}{k}
}{ \vi{x}{k+1} - \vi{\bar x}{k+1}}
\\
&\overset{CS}{\leq}
\|\nabla f(\vi{x}{k+1})
-
\nabla f(\vi{x}{k})
\|_W^* \| \vi{x}{k+1} - \vi{\bar x}{k+1}\|_W
\\&\quad+ \ve{
-
\tfrac1{\vi{\omega}{k}}
W (\vi{x}{k+1}-
 \vi{x}{k}) +
    \vi{z}{k}
}{ \vi{x}{k+1} - \vi{\bar x}{k+1}}
\\
&\overset{\eqref{eq:globalLipConstant}}{
\leq }
L_f^W
\|\vi{x}{k+1}
-
\vi{x}{k}
\|_W \| \vi{x}{k+1} - \vi{\bar x}{k+1}\|_W
\\&\qquad +
\ve{
-
\tfrac1{\bar  \omega }
W (\vi{x}{k+1}-
 \vi{x}{k})
}{ \vi{x}{k+1} - \vi{\bar x}{k+1}}
+
\ve{
    \vi{z}{k}
}{ \vi{x}{k+1} - \vi{\bar x}{k+1}}
\\
&\overset{CS}{\leq}
L_f^W
\|\vi{x}{k+1}
-
\vi{x}{k}
\|_W \| \vi{x}{k+1} - \vi{\bar x}{k+1}\|_W]
\\&\qquad +
\tfrac1{\bar  \omega }
\| W (\vi{x}{k+1}-
 \vi{x}{k})
\|_W^* \|\vi{x}{k+1} - \vi{\bar x}{k+1}\|_W
+
\|   \vi{z}{k}\|_W^*
\|\vi{x}{k+1} - \vi{\bar x}{k+1}\|_W
\\
&=
\left(
(L_f^W+\tfrac1{\bar  \omega })
\|\vi{x}{k+1}
-
\vi{x}{k}
\|_W
+
\|   \vi{z}{k}\|_W^*
\right)
\| \vi{x}{k+1} - \vi{\bar x}{k+1}\|_W
\\
&\overset{\eqref{eq:WSC}}{\leq}
\left(
(L_f^W+\tfrac1{\bar  \omega })
\|\vi{x}{k+1}
-
\vi{x}{k}
\|_W
+
\|   \vi{z}{k}\|_W^*
\right)
\sqrt{
\frac1{\kappa_f}
  ( f(\vi{x}{k+1}) - f(\vi{\bar x}{k+1}) )
}.  \tagthis\label{eavkfjfvkeavfaeva}
\end{align*}
Therefore, we can conclude that
\begin{align*}
 f(\vi{x}{k+1}) - f^*
&\overset{\eqref{eavkfjfvkeavfaeva}}{\leq}
\frac1{\kappa_f}
\left(
(L_f^W+\tfrac1{\bar  \omega })
\|\vi{x}{k+1}
-
\vi{x}{k}
\|_W
+
\|   \vi{z}{k}\|_W^*
\right)^2.
\tagthis \label{asdfjasfvjsakfa}
\end{align*}
Taking the expectation of \eqref{asdfjasfvjsakfa} with respect to the random vector $\vi{z}{k}$, we obtain
 \begin{align*}
 \Exp[f(\vi{x}{k+1}) - f(\vi{\bar x}{k+1})]
&\overset{\eqref{asdfjasfvjsakfa}}{\leq}
\frac1{\kappa_f}
\Exp\left[\left(
(L_f^W+\tfrac1{\bar  \omega })
\|\vi{x}{k+1}
-
\vi{x}{k}
\|_W
+
\|   \vi{z}{k}\|_W^*
\right)^2\right]
\\
&\leq
\frac2{\kappa_f}
\left(
(L_f^W+\tfrac1{\bar  \omega })^2
\Exp[\|\vi{x}{k+1}
-
\vi{x}{k}
\|_W^2 ]
+
\Exp[(\|   \vi{z}{k}\|_W^*
)^2]\right)
\\
&\overset{\eqref{eq:RFDM-2}}{\leq}
\frac2{\kappa_f}
\left(
(L_f^W+\tfrac1{\bar  \omega })^2
+
\beta^2 \right)   \Exp[\|\vi{x}{k}
    -\vi{x}{k+1}\|_W^2]
\\
&\overset{\eqref{eq:RFDM-3}}{\leq}
\frac2{\kappa_f}
\left(
(L_f^W+\tfrac1{\bar  \omega })^2
+
\beta^2 \right)
\frac1{\zeta}  \left( f(\vi{x}{k})  - \Exp[f(\vi{x}{k+1})] \right)
\\
&=
\underbrace{\frac2{\kappa_f}
\left(
(L_f^W+\tfrac1{\bar  \omega })^2
+
\beta^2 \right)
\frac1{\zeta}}_{c}  \left(  f(\vi{x}{k})  -f(\vi{\bar x}{k}) + \Exp[f(\vi{\bar x}{k+1})] - \Exp[f(\vi{x}{k+1})] \right).
\tagthis \label{akfskmkfasdadalwafcwa}
\end{align*}
Finally, from \eqref{akfskmkfasdadalwafcwa} we obtain that
 \begin{align*}
 \Exp[f(\vi{x}{k+1}) - f^*]
 =
 \Exp[f(\vi{x}{k+1}) - f(\vi{\bar x}{k+1})]
 \leq
\frac{c}{1+c}  \left(  f(\vi{x}{k})  -f(\vi{\bar x}{k+1}) \right)
=
\frac{c}{1+c}  \left(  f(\vi{x}{k})  -f^*  \right),
\end{align*}
and the result follows.

\subsection{Proof of Theorem \ref{thm:LC:RC-FDM}
if $\vi{z}{k} =  0$ }

Let us define an auxiliary
vector $\tilde x$
such that
\begin{align*}
\vc{\tilde x}{i}
 &= \vsubset{\proj{X}{
 \vi{x}{k} - \vi{\omega}{k}
  W^{-1}
  \vsubset{(\nabla f(\vi{x}{k})
   - \vi{z}{k})}{i}
  }}{i}.
  \tagthis \label{savejflcjesaclavc}
\end{align*}
Then we can see
that if  coordinate $i$ is chosen during iteration $k$ in Algorithm \ref{alg:sdca}
then
\begin{equation}
\vc{\vi{x}{k+1}}{j}
 = \begin{cases}
  \vc{\vi{x}{k}}{j},
  &\mbox{if }\ j\neq i,\\
 \vc{\tilde x}{i},&\mbox{otherwise.}
 \end{cases}\label{2erq243rffa}
\end{equation}
Therefore, let us estimate the expected value of $f$ at a random point $\vi{x}{k+1}$,
where the expectation is taken with respect to the selection of coordinate $i$ at iteration k.
Let $h\in \R^n$.
Then
if $\frac1{\omega_k} \geq
\max_{i} \frac{L_i}{w_i}$
we have
\begin{align*}
\Exp[f(\vi{x}{k}+\vsubset{h}{i})]
&
\overset{\eqref{eq:CWLC}}{\leq }
f(\vi{x}{k})
 + \Exp\left[ \ve{\nabla f(\vi{x}{k})}{  \vsubset{h}{i}}
 +\frac{L_i}{2 w_i} \| \vsubset{h}{i}\|^2_W\right]
 \\
&\leq
f(\vi{x}{k})
 + \Exp\left[ \ve{\nabla f(\vi{x}{k})}{  \vsubset{h}{i}}
 +\frac{1}{2\omega_k} \| \vsubset{h}{i}\|^2_W\right]
\\
&\overset{\eqref{2erq243rffa}}{=}
f(\vi{x}{k})
 + \frac1n \left(
  \ve{\nabla f(\vi{x}{k})}{ h }
 +\frac{1}{2 \omega_k} \| h\|^2_W \right)
 \\
 &=
 \frac{n-1}{n}f(\vi{x}{k})
 + \frac1n \left(
 \underbrace{
 f(\vi{x}{k})+
  \ve{\nabla f(\vi{x}{k}) - \vi{z}{k}}{ h}
 +\frac{1}{2 \omega_k} \| h\|^2_W
 }_{\mathcal{H}(h; \vi{x}{k}, \vi{z}{k})}
+
 \ve{\vi{z}{k}}{h}
 \right).
 \tagthis \label{afwwe3rwg3ag3}
\end{align*}
Now, observe that
\begin{align*}
\tilde x
&=\vi{x}{k} + \arg\min _{h : x+\vi{x}{k} \in X}  \mathcal{H}(h; \vi{x}{k}, \vi{z}{k})
\\
&=\vi{x}{k} + \arg\min _{h \in \R^n   }
\left\{ \mathcal{H}(h; \vi{x}{k}, \vi{z}{k}) + \Phi_X (x+\vi{x}{k})
\right\} =: \vi{x}{k} + \hat h,
\tagthis \label{afiowjogj3242}
\end{align*}
where $\Phi_X(x)$ is the indicator function
for the set $X$, i.e.
\begin{equation}\label{IndicatorFunction}
 \Phi_X(x) =
\begin{cases}
 0, &\mbox{if}\ x \in X,
 \\
 +\infty,&\mbox{otherwise.}
 \end{cases}
\end{equation}
From the first order optimality conditions of
\eqref{afiowjogj3242}
we have
\begin{equation}\label{eqavrg34332}
 \nabla f(\vi{x}{k}) - \vi{z}{k}
  + \frac1{\omega_k}
  W \hat h + s = 0,
\end{equation}
where $s \in \partial \Phi (\vi{x}{k} + \hat h)$.
We can define a composite gradient mapping \cite{lu2013complexity,nesterov2007gradient,
tappenden2015complexity} as
\begin{equation}
g :=  -\frac1{\omega_k} W \hat h.
\end{equation}
 Therefore, we
can observe that
\begin{equation}
  -\nabla f(\vi{x}{k})
  + \vi{z}{k}
  + g \overset{\eqref{eqavrg34332}}{\in}
 \partial \Phi (\vi{x}{k} + \hat h).
 \label{af4334t2r213}
\end{equation}
It is also easy to show that
\begin{align*}
\| \hat h\|_W^2
&=
\|\omega_k W^{-1} g \|_W^2
= \omega_k^2 (\|g\|_W^*)^2
\label{eq:awiejfoew}\tagthis
\end{align*}
and
\begin{align*}
\ve{g}{\hat h}
 = - \frac1{\omega_k} \|\hat h\|_W^2
 \overset{\eqref{eq:awiejfoew}}{=}
 - \omega_k  (\|g\|_W^*)^2.
 \tagthis \label{ara2try53tgae}
\end{align*}
Finally note that for any $y \in X$ we have
\begin{align*}
\|\vi{x}{k} + \hat h - y \|_W^2
&= \|\vi{x}{k} - y\|_W^2
+ 2 \omega_k\ve{g}{y-\vi{x}{k}}
+ \|\hat h\|_W^2
\\
&\overset{\eqref{eq:awiejfoew}}{=} \|\vi{x}{k} - y\|_W^2
+ 2 \omega_k\ve{g}{y-\vi{x}{k}}
+ \omega_k^2 (\|g\|_W^*)^2.
\tagthis \label{faarwga3wgr3gra}
\end{align*}
Now, we are ready to bound
$\mathcal{H}(h; \vi{x}{k}, \vi{z}{k})
+\Phi(x+h)$
for $h=\hat h$.
We have
\begin{align*}
&\mathcal{H}(\hat h; \vi{x}{k}, \vi{z}{k})
+\Phi(\vi{x}{k}+\hat h)\\
&=
f(\vi{x}{k})+
 \ve{\nabla f(\vi{x}{k}) - \vi{z}{k}}{\hat h}
 +\frac{1}{2 \omega_k} \|\hat h\|^2_W
 +\Phi(\vi{x}{k}+\hat h)
\\
&\overset{\eqref{af4334t2r213}}{\leq}
f(y) + \ve{\nabla f(\vi{x}{k})}{\vi{x}{k}-y}
+
 \ve{\nabla f(\vi{x}{k}) - \vi{z}{k}}{\hat h}
 +\frac{1}{2 \omega_k} \|\hat h\|^2_W
\\&\qquad +\Phi(y)
+ \ve{-\nabla f(\vi{x}{k})
  + \vi{z}{k}
  + g}{\vi{x}{k}+\hat h - y}
\\
&=
f(y)+\Phi(y)
 +\frac{1}{2 \omega_k} \|\hat h\|^2_W
+ \ve{  g}
  {\vi{x}{k}  - y}
+ \ve{  \vi{z}{k}  }
  {\vi{x}{k}  - y}
+ \ve{ g}
  {\hat  h}
\\
&\overset{\eqref{ara2try53tgae},\eqref{eq:awiejfoew}}{=}
f(y)+\Phi(y)
 +\frac{1}{2  } \omega_k  (\|g\|_W^*)^2
+ \ve{  g}
  {\vi{x}{k}  - y}
+ \ve{  \vi{z}{k}  }
  {\vi{x}{k}  - y}
- \omega_k  (\|g\|_W^*)^2
\\
&=
f(y)+\Phi(y)
 -\frac{1}{2  } \omega_k  (\|g\|_W^*)^2
+ \ve{  g}
  {\vi{x}{k}  - y}
+ \ve{  \vi{z}{k}  }
  {\vi{x}{k}  - y}
\\
&\overset{\eqref{faarwga3wgr3gra}}{=}
f(y)+\Phi(y)
  -\frac1{2 \omega_k}
 \left(
\|\vi{x}{k} + \hat h - y \|_W^2
-\|\vi{x}{k} - y\|_W^2
\right)
+ \ve{  \vi{z}{k}  }
  {\vi{x}{k}  - y}
\\
&\overset{\eqref{2erq243rffa},\eqref{afiowjogj3242}}{=}
f(y)+\Phi(y)
  -\frac1{2 \omega_k}
 \left(
n\Exp[\|\vi{x}{k+1}   - y \|_W^2]
-n \| \vi{x}{k} - y \|_W^2
\right)
+ \ve{  \vi{z}{k}  }
  {\vi{x}{k}  - y}.
\end{align*}
Now, from
 \eqref{afwwe3rwg3ag3}
 we conclude that $\forall y$ we have
\begin{align*}
\Exp[f(\vi{x}{k+1})]
 &\leq
 \frac{n-1}{n}f(\vi{x}{k})
 + \frac1n \left(
 f(y)+\Phi(y)
  -\frac n{2 \omega_k}
 \left(
\Exp[\|\vi{x}{k+1}   - y \|_W^2]
-\|\vi{x}{k} - y\|_W^2
\right)
+ \ve{  \vi{z}{k}  }
  {\vi{x}{k} +\hat h - y}
 \right),
\end{align*}
which can be equivalently written as
\begin{align*}
\Exp\left[ f(\vi{x}{k+1})
+\frac1{2\omega_k}
 \|\vi{x}{k+1}   - y \|_W^2\right]
 &\leq
 f(\vi{x}{k})
 +\frac 1{2 \omega_k}
\|\vi{x}{k} - y\|_W^2
 -
 \frac{1}{n}
 (f(\vi{x}{k})-
 f(y)-\Phi(y))
+
\frac1n \ve{  \vi{z}{k}  }
  {\vi{x}{k} +\hat h - y}.
\end{align*}
If we choose $y = \vi{\bar x}{k}$ then
the latter inequality reads as follows:
\begin{align*}
\Exp\left[ f(\vi{x}{k+1})
+\frac1{2\omega_k}
 \|\vi{x}{k+1}   - \vi{\bar x}{k} \|_W^2\right]
 &\leq
 f(\vi{x}{k})
 +\frac 1{2 \omega_k}
\|\vi{x}{k} - \vi{\bar x}{k}\|_W^2
 -
 \frac{1}{n}
 (f(\vi{x}{k})-
 f^* )
+
\frac1n \ve{  \vi{z}{k}  }
  {\vi{x}{k} +\hat h - \vi{\bar x}{k}}.
\end{align*}
From the definition of $\bar x$ we obtain that
$\|\vi{x}{k+1} - \vi{\bar x}{k+1}\|_W
 \leq
\|\vi{x}{k+1} - \vi{\bar x}{k}\|_W
$
and therefore
\begin{align*}
\Exp\left[ f(\vi{x}{k+1})-f^*
+\frac1{2\omega_k}
 \|\vi{x}{k+1}   - \vi{\bar x}{k+1} \|_W^2\right]
 &\leq
 (1-\tfrac1n)
 (f(\vi{x}{k})-f^*)
 +\frac 1{2 \omega_k}
\|\vi{x}{k} - \vi{\bar x}{k}\|_W^2
+
\frac1n \ve{  \vi{z}{k}  }
  {\vi{x}{k} +\hat h - \vi{\bar x}{k}}.
\end{align*}
Let us assume that $\forall k: \vi{z}{k} = 0$.
Then let us define
$c = \frac{2 \bar \omega \kappa}{n(2 \bar \omega \kappa+1)} \in (0,1)$. Then
 \begin{align*}
\Exp\left[ f(\vi{x}{k+1})-f^*
+\frac1{2\omega_k}
 \|\vi{x}{k+1}   - \vi{\bar x}{k+1} \|_W^2\right]
 &\leq
 (1-c)
 \left(
 f(\vi{x}{k})-f^*
 +\frac 1{2 \bar \omega}
\|\vi{x}{k} - \vi{\bar x}{k}\|_W^2
\right). \tagthis \label{Afw43tr2a3t34}
\end{align*}
Therefore
$$
\Exp[f(\vi{x}{k})-f^*]
\leq
\Exp\left[ f(\vi{x}{k})-f^*
+\frac1{2\omega_k}
 \|\vi{x}{k}   - \vi{\bar x}{k} \|_W^2\right]
 \overset{\eqref{Afw43tr2a3t34}}{\leq}
 (1-c)^k \left(f(\vi{x}{0})-f^*
 +\frac1{2\bar \omega} \|\vi{x}{0} - \vi{\bar x}{0}\|_W^2\right).
$$

\subsection{Proof of Theorem \ref{thm:LC:RC-FDM}
if $\vi{z}{k}\neq 0$}

The proof follows similar arguments to the proof of Theorem \ref{thm:LC:RC-FDM}
when $\vi{z}{k}= 0$. Let us define an auxiliary
vector $\tilde x$ in the same way as in
\eqref{savejflcjesaclavc}. Then we can see
that if  coordinate $i$ is chosen during iteration $k$ in Algorithm \ref{alg:sdca}
then \eqref{2erq243rffa} holds.
Therefore, let us estimate
the expected value of
$f$ at a random point $\vi{x}{k+1}$,
where the expectation is taken with respect to the selection of coordinate $i$ at iteration k.
Let $h\in \R^n$.
Then if $\frac1{\omega_k} \geq \max_{i} \frac{L_i}{w_i}$
we have that \eqref{afwwe3rwg3ag3} holds.
Now, observe that
\begin{align*}
\tilde x
&=\vi{x}{k} + \arg\min _{h : x+\vi{x}{k} \in X}  \mathcal{H}(h; \vi{x}{k}, \vi{z}{k})
\\
&=\vi{x}{k} + \arg\min _{h \in \R^n   }
\left\{ \mathcal{H}(h; \vi{x}{k}, \vi{z}{k}) + \Phi_X (h+\vi{x}{k})
\right\} =: \vi{x}{k} + \hat h,
\tagthis \label{afiowjogj3242}
\end{align*}
where $\Phi_X(x)$ is indicator function
for set $X$, \eqref{IndicatorFunction}. 
Now, we have
\begin{align*}
\mathcal{H}(\hat h; \vi{x}{k}, \vi{z}{k})
&=
\min _{h \in \R^n   }
\left\{  f(\vi{x}{k})+
  \ve{\nabla f(\vi{x}{k}) - \vi{z}{k}}{ h}
 +\frac{1}{2 \omega_k} \| h\|^2_W + \Phi_X (h+\vi{x}{k})
\right\}
\\
&=
\min _{y \in \R^n   }
\left\{  f(\vi{x}{k})+
  \ve{\nabla f(\vi{x}{k}) - \vi{z}{k}}{ y-\vi{x}{k}}
 +\frac{1}{2 \omega_k} \| y-\vi{x}{k}\|^2_W + \Phi_X (y)
\right\}
\\
&\leq
\min _{\lambda  \in [0,1]   }
\left\{  f(\lambda \vi{\bar x}{k} + (1-\lambda)\vi{x}{k})+
  \ve{ - \vi{z}{k}}{
  \lambda (\vi{\bar x}{k}- \vi{x}{k} )}
 +\frac{1}{2 \omega_k} \| \lambda (\vi{\bar x}{k}- \vi{x}{k} )\|^2_W + \Phi_X (\lambda (\vi{\bar x}{k}- \vi{x}{k} )+\vi{x}{k})
\right\}
\\
&\leq
\min _{\lambda  \in [0,1]   }
\left\{ \lambda f(\vi{\bar x}{k})
+(1-\lambda) f(\vi{x}{k})
+
\lambda   \|\vi{z}{k}\|_W^*
  \| \vi{\bar x}{k}- \vi{x}{k} \|_W
 +\frac{\lambda^2}{2 \omega_k} \|  \vi{\bar x}{k}- \vi{x}{k}  \|^2_W
\right\}.
\end{align*}
Note that from
\eqref{2erq243rffa}
and \eqref{afiowjogj3242}
we have
\begin{equation}
\|\hat h\|_W^2
 = \sum_{i=1}^n \| \vsubset{\hat h}{i} \|_W^2
 =n \Exp [ \| \vi{x}{k+1} - \vi{x}{k} \|_W^2 ]
 \overset{\eqref{eq:RCFDM-3}}{\leq}
 \frac{n}{\zeta}
 \Exp [  f(\vi{x}{k}) - f(\vi{x}{k+1}) ].
 \label{adfewfrvfsafda}
\end{equation}
Therefore, we conclude that
\begin{align*}
\Exp[f(\vi{x}{k+1})
-f^* ]
&
\overset{\eqref{afwwe3rwg3ag3},\eqref{eq:RCFDM-2}}{\leq }
  \min _{\lambda  \in [0,1]   }
\{
 f(\vi{x}{k})-f^*
 + \tfrac1n  (
 \lambda (f(\vi{\bar x}{k})
 - f(\vi{x}{k}))
+
\lambda    \|\vi{z}{k}\|_W^*
  \| \vi{\bar x}{k}- \vi{x}{k} \|_W
\\&\qquad +\frac{\lambda^2}{2 \omega_k} \|  \vi{\bar x}{k}- \vi{x}{k}  \|^2_W
+
 \|\vi{z}{k}\|_W^* \|\hat h\|_W
 )
\}
\\
&
\overset{\eqref{eq:WSC}}{\leq }
  \min _{\lambda  \in [0,1]   }
\{
 f(\vi{x}{k})-f^*
 + \tfrac1n  (
 -
 \lambda (
   f(\vi{x}{k})-f^*)
+
\lambda    \|\vi{z}{k}\|_W^*
  \| \vi{\bar x}{k}- \vi{x}{k} \|_W
\\&\qquad
+\frac{\lambda^2}{2 \omega_k \kappa_f} (f(\vi{x}{k})-f^*)
+
 \|\vi{z}{k}\|_W^* \|\hat h\|_W
 )
\}.
\end{align*}
Now, let us denote by
$\xi_k = f(\vi{x}{k})-f^*$
and
$\xi_{k+1} = \Exp[f(\vi{x}{k})-f^*]$
(where the expectation is with respect to the random choice $i$ during the $k$-th iteration).
Notice that
\begin{equation}\label{avfefvsvsafcas}
(\|\vi{z}{k}\|_W^*)^2
=
\sum_{i=1}^n
(\|\vsubset{(\vi{z}{k})}{i}\|_W^*)^2
\overset{\eqref{eq:RCFDM-2},\eqref{eq:RCFDM-3}}{\leq}
n \frac{\beta^2}{\zeta} ( \vi{\xi}{k}-\vi{\xi}{k+1}).
\end{equation}
Therefore we have
\begin{align*}
\vi{\xi}{k+1}
&\leq
  \min _{\lambda  \in [0,1]   }
\{
 \xi_k
 + \tfrac1n  (
 -
 \lambda \xi_k
+
\lambda    \|\vi{z}{k}\|_W^*
  \| \vi{\bar x}{k}- \vi{x}{k} \|_W
+\frac{\lambda^2}{2 \omega_k \kappa_f} \vi{\xi}{k}
+
 \|\vi{z}{k}\|_W^* \|\hat h\|_W
 )
\}
\\
&\overset{\eqref{avfefvsvsafcas},\eqref{adfewfrvfsafda}}{\leq}
 \min _{\lambda  \in [0,1]   }
\{
 \vi{\xi}{k}
 + \tfrac1n  (
 -
 \lambda \vi{\xi}{k}
+
\lambda    \|\vi{z}{k}\|_W^*
  \| \vi{\bar x}{k}- \vi{x}{k} \|_W
+\frac{\lambda^2}{2 \omega_k \kappa_f} \vi{\xi}{k}
+
\frac{n \beta}{\zeta}
 (\vi{\xi}{k}-\vi{\xi}{k+1}))
\}
\end{align*}
which is equivalent to

\begin{align*}
(1+\tfrac{\beta}{\zeta})
\vi{\xi}{k+1}
&{\leq}
(1+\tfrac{\beta}{\zeta}) \vi{\xi}{k}
+ \min _{\lambda  \in [0,1]   }
\{
 -
 \tfrac1n  \lambda \vi{\xi}{k}
+
\tfrac1n   \lambda    \|\vi{z}{k}\|_W^*
  \| \vi{\bar x}{k}- \vi{x}{k} \|_W
+\tfrac1n  \tfrac{\lambda^2}{2 \omega_k \kappa_f} \vi{\xi}{k}
\}
\\
&
\overset{\eqref{avfefvsvsafcas},\eqref{eq:WSC}
}{\leq}
(1+\tfrac{\beta}{\zeta}) \vi{\xi}{k}
+ \min _{\lambda  \in [0,1]   }
\{
 -
 \tfrac1n  \lambda \vi{\xi}{k}
+
\tfrac1n   \lambda
\sqrt{n \tfrac{\beta^2}{\zeta} ( \vi{\xi}{k}-\vi{\xi}{k+1})}
 \sqrt{\tfrac1{\kappa_f} \vi{\xi}{k}}
+\tfrac1n  \tfrac{\lambda^2}{2 \omega_k \kappa_f} \vi{\xi}{k}
\}.
\end{align*}
Using the fact that
$\forall a,b\in \R_+$ we have
$\sqrt{a b}\leq \frac12a + \frac12b$ we obtain that
\begin{align*}
(1+\tfrac{\beta}{\zeta})
\vi{\xi}{k+1}
&{\leq}
(1+\tfrac{\beta}{\zeta}) \vi{\xi}{k}
+ \min _{\lambda  \in [0,1]   }
\{
 -
 \tfrac1n  \lambda \vi{\xi}{k}
+
\sqrt{   \tfrac{ \beta^2}{\zeta} ( \vi{\xi}{k}-\vi{\xi}{k+1})}
 \sqrt{\tfrac{\lambda^2}{n}\tfrac1{ \kappa_f} \vi{\xi}{k}}
+\tfrac1n  \tfrac{\lambda^2}{2 \omega_k \kappa_f} \vi{\xi}{k}
\}
\\
&{\leq}
(1+\tfrac{\beta}{\zeta}) \vi{\xi}{k}
+ \min _{\lambda  \in [0,1]   }
\{
 -
 \tfrac1n  \lambda \vi{\xi}{k}
+   {   \frac{ \beta^2}{2\zeta} ( \vi{\xi}{k}-\vi{\xi}{k+1})}
+\frac12 {\frac{\lambda^2}{n}\frac1{ \kappa_f} \vi{\xi}{k}}
+\tfrac1n  \tfrac{\lambda^2}{2 \omega_k \kappa_f} \vi{\xi}{k}
\}.
\end{align*}
Therefore, we obtain
\begin{align}\label{xiforzneq0}
(1+\tfrac{\beta}{\zeta}+\tfrac{ \beta^2}{2\zeta} )
\vi{\xi}{k+1}
&{\leq}
(1+\tfrac{\beta}{\zeta} +\tfrac{ \beta^2}{2\zeta} ) \vi{\xi}{k}
+\frac{1}{n \bar \omega  \kappa_f} \min _{\lambda  \in [0,1]   }
\{
 -
   \lambda   \bar \omega \kappa_f
+\frac{\lambda^2}2 (1+\bar \omega )
\}
\vi{\xi}{k}.
\end{align}
The optimal $\lambda^*$ that minimizes the above expression is
$$
    \lambda^* = \max\left\{1,  \frac{\bar \omega  \kappa_f}{\bar \omega  +1} \right\}.
$$
Consider now two cases:
\begin{itemize}
\item
$\lambda^*< 1$.
In this case
$$
 -
   \lambda^*   \bar \omega \kappa_f
+\frac{(\lambda^*)^2}2 (1+\bar \omega )
 =
 -\frac12
   \frac{(\bar \omega  \kappa_f)^2}{\bar \omega  +1} .  $$
   Combining this with \eqref{xiforzneq0} gives
   \begin{align*}
(1+\tfrac{\beta}{\zeta}+\tfrac{ \beta^2}{2\zeta} )
\vi{\xi}{k+1}
&{\leq}
(1+\tfrac{\beta}{\zeta} +\tfrac{ \beta^2}{2\zeta}-\frac1{2n}
   \frac{ \bar \omega  \kappa_f }{\bar \omega  +1}  )
\vi{\xi}{k},
\end{align*}
which is equivalent to
   \begin{align*}
\vi{\xi}{k+1}
&{\leq}
\left(1-\frac1{2n}
   \frac{ \bar \omega  \kappa_f }{\bar \omega  +1}
   \frac{2\zeta}{2\zeta+ 2\beta  + \beta }  \right)
\vi{\xi}{k}.
\end{align*}
\item $\lambda^*=1$.
In this case $
\frac{\bar \omega \kappa_f}{\bar \omega+1} \geq 1$
and hence
$$
 -
   \lambda^*   \bar \omega \kappa_f
+\frac{(\lambda^*)^2}2 (1+\bar \omega )
 =
 -
       \bar \omega \kappa_f
+\frac12 (1+\bar \omega )
\leq
 -
       \bar \omega \kappa_f
+\frac12  \bar \omega \kappa_f
=-\frac12 \bar \omega \kappa_f. $$
Therefore, from \eqref{xiforzneq0} we can conclude that
\begin{align*}
\vi{\xi}{k+1}
&{\leq}
\left(1-
\frac{ \zeta}{n(2\zeta+ 2 \beta + \beta^2)  } \right) \vi{\xi}{k}.
\end{align*}
\end{itemize}

\end{document}